# Electrodermal Activity as a Unimodal Signal for Aerobic Exercise Detection in Wearable Sensors

1st Rena Krishna
Odle Middle School
ImagineQ Labs
Bellevue, WA, USA
ORCID 0009-0005-7145-491X

2nd Ramya Sankar

ImagineQ Labs
Bellevue, WA, USA

3rd Shadi Ghiasi

Cambridge Center for International
Research, United Kingdom

*Abstract*—Electrodermal Activity (EDA) is a non-invasive physiological signal widely available in wearable devices and reflects sympathetic nervous system (SNS) activation. Prior multi-modal studies have demonstrated robust performance in distinguishing stress and exercise states when EDA is combined with complementary signals such as heart rate and accelerometry. However, the ability of EDA to independently distinguish sustained aerobic exercise from low-arousal states under subject-independent evaluation remains insufficiently characterized.

This study investigates whether features derived exclusively from EDA can reliably differentiate rest from sustained aerobic exercise. Using a publicly available dataset collected from thirty healthy individuals, EDA features were evaluated using benchmark machine learning models with leave-one-subject-out (LOSO) validation.

Across models, EDA-only classifiers achieved moderate subject-independent performance, with phasic temporal dynamics and event timing contributing to class separation. Rather than proposing EDA as a replacement for multimodal sensing, this work provides a conservative benchmark of the discriminative power of EDA alone and clarifies its role as a unimodal input for wearable activity-state inference.

## I. INTRODUCTION

Aerobic exercise is a physiological state characterized by sustained cardiovascular, respiratory, and metabolic regulation that supports continuous adenosine triphosphate (ATP) production through mitochondrial oxidative phosphorylation [1], [2]. It is typically performed at moderate intensities at or below the ventilatory threshold (VT), where oxygen supply is sufficient to meet muscular metabolic demand, and is accompanied by elevated heart rate, ventilation, and muscle perfusion [3]. During aerobic exercise, skeletal muscle glucose uptake increases driven by both insulin-dependent and insulin-independent pathways, resulting in a net reduction in circulating blood glucose levels [4], [5]. This metabolic profile distinguishes aerobic exercise from anaerobic exercise and stress, which elicit acute sympathetic responses without sustained glucose utilization. In aerobic exercise, physical arousal is regulated and proportional to workload, whereas psychological stress induces rapid and often disproportionate arousal, and anaerobic exercise elicits brief, high-intensity sympathetic responses. Accurate identification of aerobic exercise is therefore relevant for wearable health monitoring systems and for applications in which prolonged physical exertion has distinct physiological or metabolic implications.

Electrodermal Activity (EDA) is a non-invasive physiological signal governed by sudomotor activation and reflects changes in skin conductance driven by eccrine sweat glands [6]. Because EDA is not directly modulated by parasympathetic input, it has been widely studied as a marker of autonomic arousal, particularly in the context of psychological stress and emotional activation [4]. Extensive prior work has demonstrated the sensitivity of EDA to changes in sympathetic tone across cognitive, affective, and physiological states, establishing it as a robust indicator of SNS activity [7], [8], [9], [10].

Recent wearable studies have increasingly adopted multimodal approaches that combine EDA with cardiovascular and motion-based signals to improve classification performance. Large-scale studies by Hongn et al., demonstrated strong discrimination among stress, aerobic, anaerobic, and rest when EDA was integrated with heart rate and accelerometry [11]. While these results highlight the effectiveness of multimodal sensing, they do not isolate the specific contribution of EDA itself. Thus, it remains unclear which aspects of activity-state discrimination arise from the sympathetic sudomotor responses versus complementary contributions from cardiovascular dynamics or movement patterns.

Other notable work by Posada-Quintero et al., has characterized how spectral shifts in EDA vary across rest, walking, and running [12]. While such studies provide important physiological insight into EDA responses to physical exertion, they do not assess whether these changes alone are sufficient to reliably classify activity states or generalize across individuals using subject-independent evaluation.

Evaluating EDA as a standalone signal therefore enables direct assessment of how reliably sympathetic sudomotor responses alone differentiate sustained aerobic exercise from low-arousal states. By analyzing EDA independently of cardiovascular and motion-based cues, this approach clarifies EDA's role within wearable sensing frameworks, especially those that prioritize low power consumption, robustness to sensor dropout, or minimal sensing configurations.

While modalities such as functional near-infrared spectroscopy (fNIRS) and electroencephalography (EEG) have been explored for related assessments, their limited portability restricts use in wearable systems [13]. In contrast, EDA can be acquired with low-power, wrist-worn sensors. Compared with wearable signals like blood volume pulse (BVP) and heart rate variability (HRV), EDA is less directly influenced by motion, hydration, or fitness-related cardiovascular adaptations. For example, fitness level does not directly modulate sympathetic sudomotor firing but substantially alters HRV and BVP though long-term cardiovascular and autonomic adaptations [8] [14].

Accordingly, the objective of this study is not to outperform multimodal activity recognition systems, but to conservatively assess whether EDA alone contains reproducible, subject-independent signatures of sustained aerobic exercise. Using a publicly available wearable dataset, this work establishes a benchmark characterization of the strengths and limitations of EDA-only models and informs real-world extensions.

## II. MATERIALS AND METHODS

### A. Dataset Description

This study utilized a publicly available, multimodal dataset of physiological signals collected from 30 subjects during a structured aerobic exercise session [15][16]. Table 2.1 summarizes the demographic characteristics of the thirty subjects that completed the aerobic protocol. From this dataset, Subjects S03 and S07 were removed due to incomplete protocol completion, and S11 due to data quality issues. The remaining 27 subjects were included in this study. It is noteworthy that the study population consisted of healthy young adults, which may limit generalization to pediatric, geriatric, or clinical populations.

TABLE 2.1

DATASET DEMOGRAPHIC CHARACTERISTICS

| Subject | Gender | Age | Height (cm) | Weight (kg) | Exercises regularly |
|---|---|---|---|---|---|
| S01 | m | 21 | 192 | 84 | Yes |
| S02 | m | 20 | 185 | 95 | No |
| S03 | m | 20 | 175 | 62 | Yes |
| S04 | m | 21 | 174 | 70 | Yes |
| S05 | m | 21 | 173 | 72 | Yes |
| S06 | m | 21 | 172 | 70 | Yes |
| S07 | m | 19 | 184 | 88 | Yes |
| S08 | m | 20 | 174 | 67 | Yes |
| S09 | m | 19 | 174 | 63 | Yes |
| S10 | m | 21 | 180 | 80 | Yes |
| S11 | m | 21 | 183 | 64 | Yes |
| S13 | m | 21 | 175 | 65 | Yes |
| S14 | m | 19 | 182 | 85 | Yes |
| S15 | m | 21 | 176 | 77 | Yes |
| S16 | m | 20 | 168 | 61 | Yes |
| S17 | m | 22 | 173 | 78 | Yes |
| S18 | m | 21 | 183 | 80 | Yes |
| f01 | f | 25 | 152 | 61 | No |
| f02 | f | 29 | 164 | 80 | No |
| f03 | f | 26 | 160 | 61 | Yes |
| f04 | f | 29 | 168 | 56 | No |
| f05 | f | 21 | 165 | 55 | Yes |
| f06 | f | 21 | 169 | 58 | No |
| f07 | f | 21 | 163 | 47 | No |
| f08 | f | 22 | 158 | 50 | No |
| f09 | f | 21 | 170 | 56 | Yes |
| f10 | f | 21 | 172 | 65 | Yes |
| f11 | f | 31 | 170 | 84 | No |
| f12 | f | 30 | 158 | 97 | No |
| f13 | f | 29 | 154 | 56 | No |

The physiological signals were collected using a wearable Empatica E4 wristband with EDA sampled at 4Hz. The aerobic exercise session followed the Storer-Davis maximal bicycle protocol, consisting of approximately 35 minutes of continuous stationary cycling [15], [17].

For each participant, the maximum resistance was first determined as the highest workload at which pedaling could not be sustained. The protocol commenced with a 3-minute warm-up during which the participants pedaled without resistance. Following the warm up, the exercise began at low resistance (20% of maximum) and progressed through three 3-minute stages at 60, 70, and 75 rpm, with resistance gradually increasing to a medium level (30% of maximum). The protocol then included four additional stages of increasing resistance: the first two lasting 3 minutes each and the second two lasting 2 minutes each. The final stage was performed at a fixed medium-high resistance (50% of maximum) and consisted of three 2-minute periods at 100, 105, and 110 rpm. Upon completion of this exercise protocol, the participants completed a 4-minute cooldown period with no resistance, and then 2 minutes of standing still [15].

From this dataset, the EDA signal was extracted for the 27 subjects, while all other physiological signals were excluded from analysis. To streamline signal processing, a windowing approach was applied [7], [18]. A 120-second rest window (0-120s) was extracted from the warm-up phase to ensure consistent sensor contact and minimize movement transition. A 180-second aerobic window (1560-1740s) was extracted, spanning the final 120 seconds of cycling at 110 rpm (medium-high resistance) and the first 60 seconds of cooldown. Aerobic segments were intentionally extended into early recovery to capture sustained sympathetic activation, which persists beyond mechanical cessation of activity in EDA. Overall, the selected windows represented stable physiological states with minimal transition effects, reducing label ambiguity and enabling conservative evaluation of EDA-based classification.

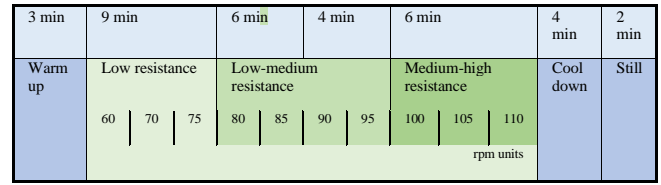

Fig. 2.1 Aerobic Exercise Protocol Stages

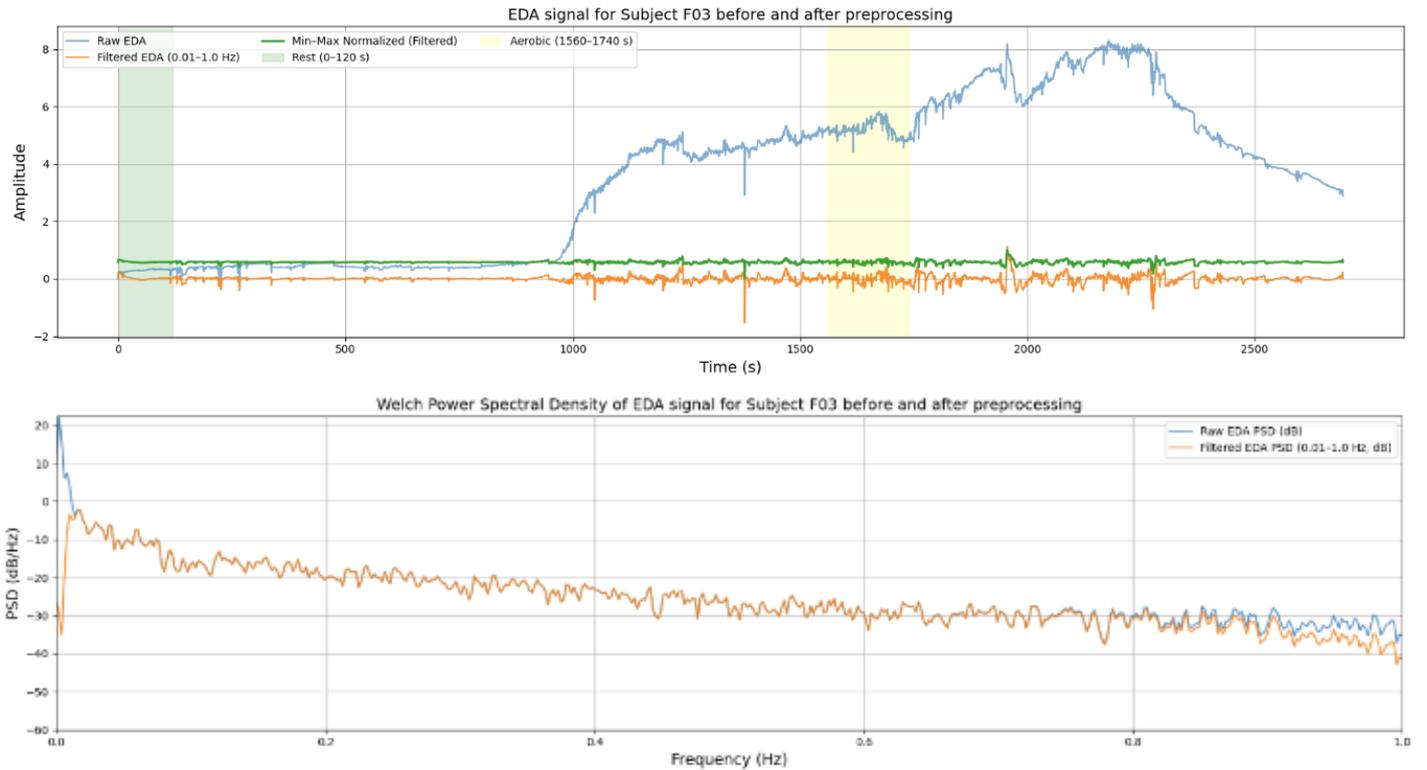

Fig. 2.2 Electrodermal activity signal before and after pre-processing

*B. Pre-processing*

The EDA signals were preprocessed to attenuate baseline drift and high frequency noise. For each subject, the full EDA recording was filtered using a zero-phase fourth-order Butterworth bandpass filter (0.01-1.0Hz) [19], where the lower cutoff attenuates slow drift, while the upper cutoff suppresses high-frequency noise and motion-related artifacts.

The band-limited signals were then normalized using min–max scaling prior to feature extraction. This normalization reduces inter-subject amplitude variability arising from differences in skin properties, sensor contact, or sweat gland density, and emphasizes relative temporal dynamics and event structure for subject-independent evaluation. All tonic (low-frequency), phasic (SCR-related), and spectral features were computed from the band-limited, normalized EDA signal, yielding scale-invariant features suitable for subject-independent evaluation.

In this study, tonic activity is operationally defined as the low-frequency component of the band-limited EDA signal obtained with a low-pass filter with a cutoff of 0.05Hz. Phasic activity is operationally defined as SCR-related dynamics within higher frequency bands (0.1-0.5Hz) of the band-limited EDA signal and characterized through event timing, recovery dynamics, and relative amplitude rather than explicit tonic subtraction.

The Butterworth filter was selected due to its maximally flat frequency response, which avoids passband and stopband ripples and preserves relative signal amplitudes [19]. This property is especially important for EDA where distortion of low-frequency dynamics can affect tonic and phasic characterization. In contrast, filtering approaches like the Savitzky-Golay perform polynomial smoothing without explicit frequency selectivity, and Chebyshev filters introduce ripple artifacts that distort EDA dynamics [20], [21]. Butterworth filtering was therefore chosen as the most conservative approach for pre-processing to isolate EDA and minimize distortion. A representative example of the signals before and after preprocessing are shown in Figure 2.2.

*C. Feature Extraction*

Following pre-processing, for each subject, two feature vectors were generated: one corresponding to the rest condition (first 120 seconds of the recording), and the other corresponding to the aerobic exercise condition (1560-1740 seconds of the recording). This resulted in two labelled observations per subject, representing rest and aerobic state respectively, using which a feature matrix was constructed at the subject level.

A total of twenty-three initial candidate features were extracted from both the time domain and the frequency domain. A three-step process was followed to extract features. Firstly, a time-series analysis of the pipeline was used to extract both statistical measures and EDA-specific signal characteristics from the time-domain. Tonic features were extracted with a low-pass filter and phasic with a secondary band-pass filter (see Pre-processing).

Secondly, frequency analysis was performed using the power spectral density of the signal, obtained using Welch's method. Welch's method estimates power spectral density by averaging windowed periodograms across overlapping signal segments, providing a robust and low variance spectral estimate. Frequency features helped capture sustained oscillatory activation associated with prolonged exertion. Welch's PSD estimate is formally represented below [22].

$$\hat{P}(f) = \frac{1}{K} \sum_{k=1}^{K} \frac{1}{fs\,U} |DFT\{x_k(n) \cdot w(n)\}|^2$$

Where:

P^(f) represents the estimated power spectral density at frequency f,

U is the window normalization factor

K is the number of segments

$x_k(n)$ denotes the kth signal segment in time domain

w(n) is a Hann window

Welch's method was selected over direct FFT-based spectral estimation due to its improved variance reduction and robustness to noise, which is desirable for low-frequency physiological signals such as EDA.

Thirdly, tonic trajectory (sequence) features were initially explored using sliding window summaries (e.g., tonic slope representing overall trend, activation duration representing how long the trend was sustained) following prior work emphasizing the benefits of segmentation for downstream systems [23]. However, these sequence-based features were excluded from the final model set because they behaved as protocol-dependent proxies under fixed-window segmentation (see Feature Selection and Limitations).

In addition to standard statistical features extracted using Python's standard scipy, pandas, biosppy libraries, custom frequency features were also developed to characterize spectral structure. Specifically, the number of local maxima or spectral peaks in the PSD was computed by identifying the frequency bins where spectral slope changed from positive to negative. Furthermore, the second-largest spectral peak amplitude was also extracted by masking the dominant peak in the PSD and computing the maximum of the remaining spectrum. These features were designed to uniquely distinguish the repeated, sustained sympathetic activation during aerobic activity and differentiate both spectral complexity and energy distribution between rest and aerobic conditions. A summary of extracted features is shared in Table 2.3.

TABLE 2.3
INITIAL CANDIDATE FEATURES

| Feature Name | Feature Type |
|---|---|
| Mean Raw EDA | Raw signal, Time-domain |
| Kurtosis | Raw signal, Time-domain |
| Skewness | Raw signal, Time-domain |
| IQR | Raw signal, Time-domain |
| Mean Tonic EDA | Tonic characteristic |
| Tonic Variance | Tonic characteristic |
| Mean Phasic EDA | Phasic characteristic |
| Phasic Variance | Phasic characteristic |
| Phasic Mean Rise Time | Phasic temporal dynamics |
| Phasic Mean Recovery Time | Phasic temporal dynamics |
| Peak density | Phasic event timing |
| SCR Peaks per min | Phasic event timing |
| SCR Interval Mean | Phasic event timing |
| SCR Interval Max | Phasic event timing |
| SCR Interval RMSSD | Phasic event timing |
| Peak Amplitude | Phasic characteristic |
| Second_Amplitude | Phasic characteristic |
| Freq Peak Amplitude | Frequency-domain structure |
| Freq 2ndPeak Amplitude | Frequency-domain structure |
| PSD band power | Frequency band power |
| Tonic mean slope | Tonic trajectory/sequence |
| Tonic ratio up | Tonic trajectory/sequence |
| Activation duration | Tonic trajectory/sequence |

*D. Feature Selection*

To reduce feature redundancy and multi-collinearity, several visualization and data mining techniques were used. After an initial data-quality validation, several features like Mean Raw EDA, Mean Phasic EDA, Tonic variance, Phasic Variance, Tonic Ratio Up and Activation Duration were removed because they exhibited near-zero or constant values, due to signal normalization and windowing constraints. These features were excluded prior to applying more sophisticated data mining techniques to prevent numerical instability or introducing a protocol-specific bias in the pipeline.

A correlation analysis was performed across all remaining features using pairwise correlation coefficients. As a result, highly correlated features such as Peak density, SCR Interval Mean, SCR Interval RMSSD, 2$^{nd}$ Amplitude, Frequency 2$^{nd}$ Amplitude, and PSD Band power were removed. Feature selection was then performed using Recursive Feature Elimination (RFE) with a logistic regression classifier as the base estimator to rank features based on their contribution to the classifier's decision boundary and their informative value [24], [25]. Prior to RFE, missing values were imputed using the median and features standardized to ensure comparability. After RFE ranking, the top eight features were taken into modeling. SCR Interval Max and Freq Peak Amplitude were removed as they were ranked lower than the other features. Tonic mean slope was initially included but eventually removed during LOSO validation due to overfitting concerns. The remaining features were used in model development.

A limitation of this approach was that RFE was applied as a global feature ranking step rather than nested within cross-validation, which may introduce an optimistic bias; accordingly reported feature rankings are indicative rather than definitive. This limitation was partially addressed by subject-independent LOSO validation during model evaluation to mitigate overfitting and assess generalization to unseen individuals. Principal Component Analysis (PCA) was not included in the final modelling pipeline to retain feature-level interpretability. The formulas for all selected features are shown in Table 2.4. A visual representation of the correlation matrix with features selected by RFE and used in model development is shown in Fig. 2.4 [24].

In addition to data mining techniques, statistical distribution analysis was performed using boxplots to assess the quality of RFE-selected features to be used in modeling. The boxplot analysis of these selected features is shown in Fig. 2.5.

TABLE 2.4
FORMULAS FOR SELECTED FEATURES

| Feature Name | Computation |
|---|---|
| Mean Tonic EDA | Mean Tonic EDA represents the statistical mean of the low frequency (0.05Hz) component of the band-pass filtered and normalized EDA signal. This metric captures the sustained low-frequency sympathetic activation patterns while remaining invariant to absolute conductance scaling across subjects [7], [12], [13], [26]. |
| Skewness | Skew measures the asymmetry in the EDA signal distribution. Positive skew indicates longer right tail while negative skew indicates longer left tail [7], [12], [17].<br><br>In EDA physiology, skewness increases when sympathetic events create occasional large positive excursions. The Scipy skew formula used in this analysis is based on the third standardized central moment:<br><br>$$\text{Skew}(x) = \frac{\mu_3}{\sigma^3}$$<br>where $\mu_3$ is the third central moment and $\sigma$ is the standard deviation. |
| Kurtosis | Kurtosis measures extreme values in distribution. A normal distribution has a kurtosis of 0, while positive kurtosis indicates heavier tails and negative kurtosis indicates lighter tails (fewer outliers) [7], [12], [17].<br><br>$$\text{Kurtosis}(\beta) = \frac{\mu_4}{\sigma^4}$$<br><br>where $\mu_4$ is the fourth central moment and $\sigma$ is the standard deviation. |
| IQR | IQR measures the spread of the **middle 50%** of values in the signal. In EDA, it reflects whether the signal is tightly clustered (low arousal) versus more dispersed (higher arousal and variability) and robust to outliers [7], [12], [14], [30].<br>$$IQR = Q_i(x) - Q_j(x)$$<br>Where i and j are the 75th and 25th percentiles, respectively. |
| Phasic Mean Rise Time | Phasic mean rise time estimates how long it takes for an SCR to rise from onset (baseline/threshold) to its peak. The onset is defined as the earliest time index before the peak $p_k$ at which the signal exceeds 50% of the peak amplitude above the baseline, using a backward search window [7], [12], [17].<br><br>$$\bar{t}_{rise} = \left\{ \frac{1}{M} \sum_{k=1}^{M} t_{rise}(p_k) \right.$$<br><br>Where $t_{rise}$ is rise time defined as:<br>$$t_{rise}(p) = \frac{p_k - o_k}{f_s}$$<br><br>Where $p_k$ is the sample index of the k-th detected phasic peak, $o_k$ is the onset index of the corresponding SCR immediately preceding $p_k$, and $f_s$ is the sampling rate. Note, that $\bar{t}_{rise} = 0$ when M=0.<br><br>Meaning: Shorter rise times indicate sharper, more rapid sympathetic responses (i.e., aerobic or stress). |
| Phasic Mean Recovery Time | Phasic mean recovery time estimates how long it takes for an SCR to decay after reaching its peak. It is defined as the time from peak until the phasic signal falls to below 50% of the peak amplitude above the baseline, within a forward search window (120s) [7], [12], [17].<br><br>For M>0, phasic mean recovery time is defined as:<br>$$\bar{t}_a = \left\{ \frac{1}{M} \sum_{k=1}^{M} t_a(p_k) \right.$$<br><br>Where $t_a$ is recovery time defined as:<br>$$t_a(p_k) = \frac{j_k - p_k}{f_s}$$<br>Where $j_k$ is the first time after $p_k$ where the signal falls below 50% of peak amplitude above the baseline. Note, that $\bar{t}_a = 0$ when M=0.<br><br>Meaning: Longer recovery times indicate more sustained sympathetic activation (i.e., aerobic or stress). |
| Peak Amplitude | Peak amplitude is the largest phasic peak amplitude in a segment, calculated using the Scipy find_peaks() function to find the largest magnitude. Peaks are detected in the phasic component using a prominence threshold set to one phasic standard deviation [1], [10], [12], [17].<br><br>Peak amplitude is mathematically equal to the maximum difference between the phasic signal peak and its baseline (mean).<br><br>$$\text{Peak Amplitude } |(P_i)| = \text{Peak}_i - \text{Onset}_i$$ |
| SCR Peaks per min | SCR peaks per minute are measures of SCR event rate calculated using the Scipy find_peaks() function and counting the number of peaks per minute where amplitude is greater than standard deviation [7], [11], [12], [14].<br>Peaks are detected in the phasic component using a prominence threshold set to one phasic standard deviation. The count is normalized by segment duration to yield peaks per minute.<br><br>SCR peaks per minute S, is defined as:<br>$$S = \frac{|P|}{T/60} = \frac{60|P|}{T}$$<br><br>Where |P| is the number of detected peaks. |

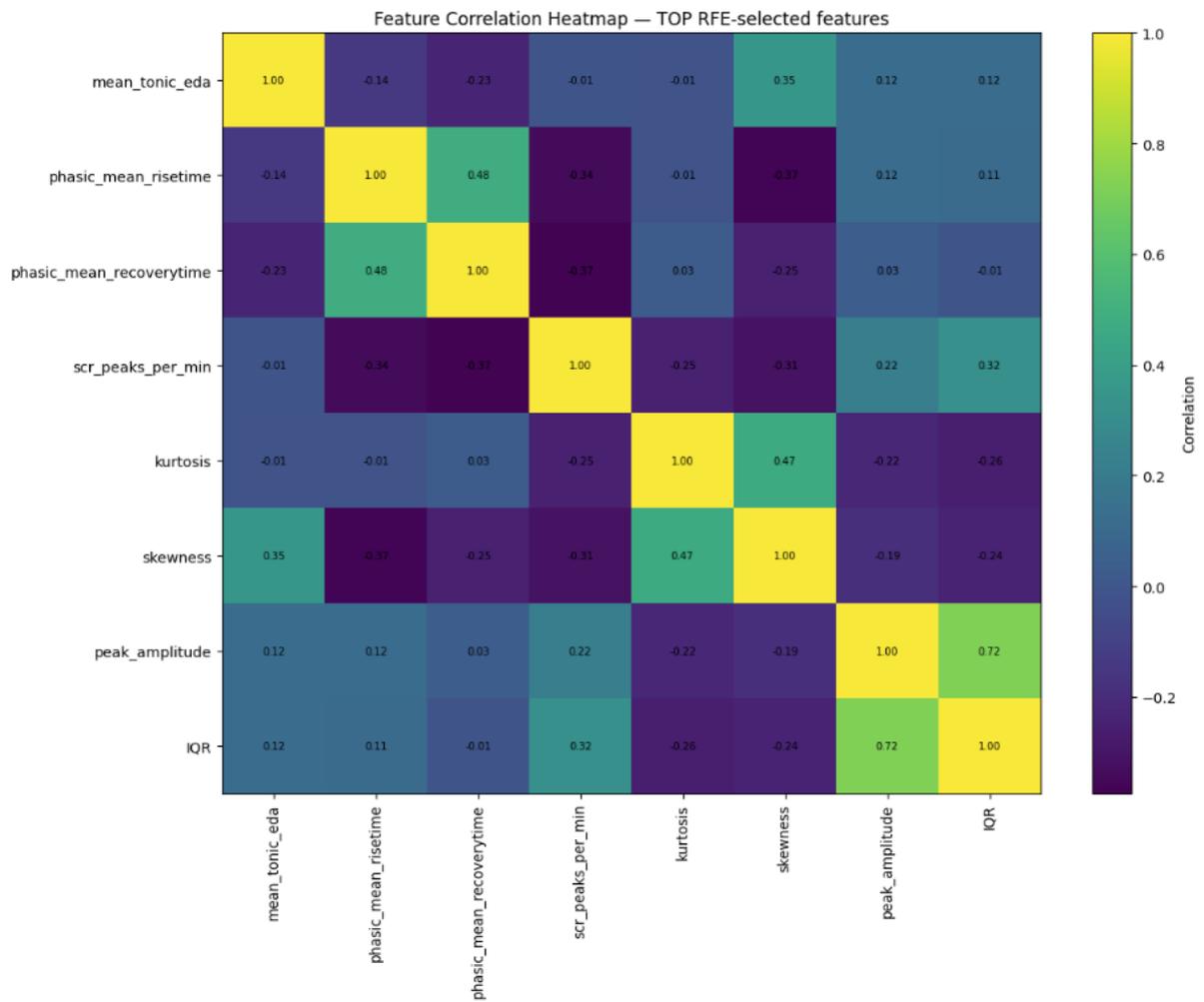

Fig. 2.4 Feature Heatmap of RFE-Selected Features used in modeling

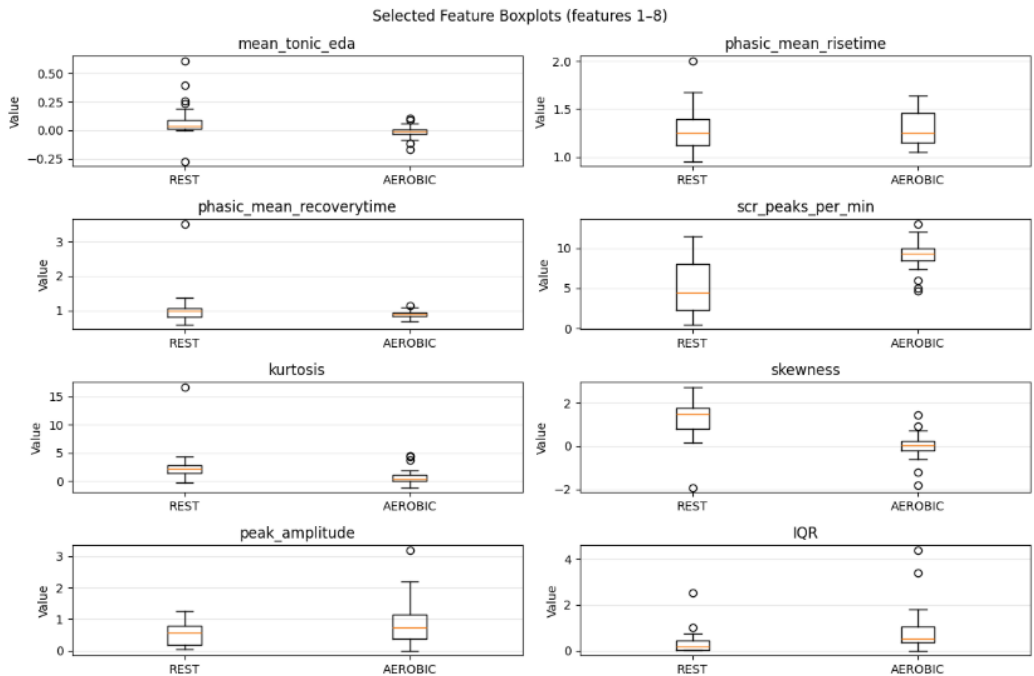

Fig. 2.5 Boxplots of RFE-selected Features used in modeling

## E. Machine Learning Model Development

To evaluate the suitability of EDA-based features for wearable and safety-critical applications, a curated set of supervised machine learning (ML) models was selected to balance interpretability, nonlinear modelling capacity, and robustness to correlated physiological features. The chosen models reflect constraints typical of wearable systems, including limited data availability, inter-subject variability, and need for generalization to unseen individuals [26].

Linear Discriminant Analysis (LDA) was employed as transparent linear baseline to assess whether rest and aerobic exercise states could be separated using simple combinations of physiologically interpretable EDA features. Gaussian Naïve Bayes (Gaussian NB) served as a probabilistic and computationally efficient baseline under the assumption of conditional feature independence. Logistic Regression (LR) was included as a low-variance, interpretable baseline model to benchmark the performance gains of more complex, nonlinear classifiers. A single Decision Tree Classifier was included as a highly interpretable baseline to examine if simple hierarchical rules could separate rest and aerobic states, while also serving as a reference for tree-based ensemble methods [27].

Additional models were evaluated like the Multi-layer Perceptron (MLP) to assess the performance of a shallow neural network architecture on engineered EDA features. Distance-based methods such as k-Nearest Neighbors (kNN) were evaluated to test whether local similarity in the engineered feature space provides reliable discrimination under subject-independent evaluation. Notably kNN demonstrated competitive performance indicating that meaningful class separation exists in the feature space despite inter-subject variability.

Support Vector Machines (SVM) with a radial basis function kernel (RBF) was selected to model nonlinear decision boundaries arising from the complex interactions between tonic, phasic, and spectral components in EDA while maintaining strong generalization on the small dataset.

Tree-based ensemble models were included due to their robustness to noise, feature correlation, and non-linear interactions commonly observed in wearable bio signals. Bagging models like Random Forest classifiers were used as a bagging ensemble to reduce variance through feature and sample averaging, providing improved generalization and stable performance across subjects, and resistance to overfitting. Extra Trees classifiers were also included as highly randomized tree ensemble to assess whether increased randomization in feature selection and split thresholds ness improves robustness and generalization in noisy, correlated feature sets [28].

Gradient-boosted decision trees were implemented to capture higher-order nonlinear dependencies with improved data efficiency. Scikit Learn's Gradient Boost Classifier was selected as a conservative and stable baseline suitable for LOSO validation. LightGBM was included as a memory-efficient gradient boosting model optimized for fast training, while XGBoost was evaluated as a high-capacity boosting model with strong regularization mechanisms and robustness to noise and correlated features [27], [28]. These models allowed trade-off assessments between boosting capacity and overfitting in small datasets.

Other models like CatBoost were considered but not included in the final model set due to performance redundancy with existing boosting models and its primary advantages for categorical features which were not present in the dataset.

## III. RESULTS

### A. Model Results and Evaluation Metrics

All models were trained and evaluated using Stratified 5-fold Cross Validation (CV) and Leave-one-subject-out (LOSO) cross-validation. Stratified 5-fold CV assessed within-dataset generalization under random splits, while LOSO assessed subject-independent performance with unseen individuals. Preprocessing steps, including imputation and scaling were performed within each training fold to prevent data leakage [7], [28].

Classification performance was quantified using accuracy, balanced accuracy, precision, recall, and F1 score. Confusion matrices were inspected to confirm balanced classification performance across both rest and aerobic classes.

Under LOSO evaluation, the models achieved moderate mean performance with substantial inter-subject variability, reflecting individual differences in EDA responses or sensor/contact effects. The LDA model performed best with an F1 score of 91% (0.91+/- 0.22) and LOSO precision of 89% (0.89 +/-0.25) and LOSO recall of 96% (0.96 +/- 0.19). Table 4.1 shows LOSO results and Fig. 4.1 shows the confusion matrix for the LOSO performance of LDA.

Under 5-fold stratified validation, kNN model with k=5 performed best achieving an F1 score of 91% (0.91 +/- 0.11) and precision of 90% (0.90 +/-0.15) and recall of 92% (0.92 +/- 0.11). Figure 4.2 shows the confusion matrix for the LDA model in 5-fold validation.

Overall, these results show that multiple model families (linear, distance-based, kernel, and ensemble methods) can achieve comparable performance when using EDA-only datasets to differentiate between rest and aerobic exercise.

TABLE 4.1
LOSO CROSS VALIDATION RESULTS

| Model | F1 Score | | Precision | | Recall | |
|---|---|---|---|---|---|---|
| | Mean | SD | Mean | SD | Mean | SD |
| LDA | 0.91 | 0.22 | 0.89 | 0.25 | 0.96 | 0.19 |
| Extra Trees | 0.89 | 0.28 | 0.87 | 0.30 | 0.93 | 0.27 |
| LR | 0.88 | 0.32 | 0.87 | 0.33 | 0.89 | 0.32 |
| kNN | 0.86 | 0.32 | 0.85 | 0.33 | 0.89 | 0.32 |
| SVM | 0.85 | 0.32 | 0.83 | 0.34 | 0.89 | 0.32 |
| MLP | 0.83 | 0.36 | 0.81 | 0.37 | 0.85 | 0.36 |
| XG Boost | 0.83 | 0.36 | 0.81 | 0.37 | 0.85 | 0.36 |
| Light GBM | 0.81 | 0.36 | 0.80 | 0.37 | 0.85 | 0.36 |
| Gaussian NB | 0.81 | 0.32 | 0.78 | 0.35 | 0.89 | 0.32 |
| Random Forest | 0.79 | 0.39 | 0.78 | 0.40 | 0.81 | 0.40 |
| Grad Boost | 0.77 | 0.39 | 0.74 | 0.40 | 0.81 | 0.40 |
| Decision Tree | 0.70 | 0.44 | 0.69 | 0.44 | 0.74 | 0.45 |

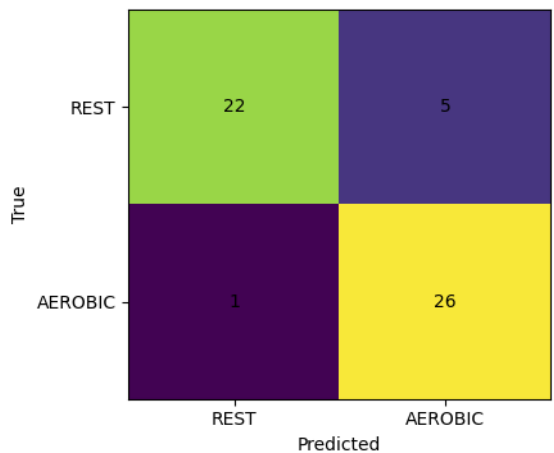

Fig. 4.1 Confusion Matrix for LDA model in LOSO validation

TABLE 4.2
5-FOLD CROSS VALIDATION RESULTS

| Model | F1 Score | | Precision | | Recall | |
|---|---|---|---|---|---|---|
| | Mean | SD | Mean | SD | Mean | SD |
| kNN | 0.91 | 0.11 | 0.90 | 0.15 | 0.92 | 0.11 |
| Extra Trees | 0.89 | 0.09 | 0.88 | 0.11 | 0.92 | 0.18 |
| LDA | 0.88 | 0.08 | 0.83 | 0.12 | 0.96 | 0.09 |
| LR | 0.86 | 0.17 | 0.90 | 0.09 | 0.88 | 0.27 |
| MLP | 0.86 | 0.18 | 0.93 | 0.10 | 0.84 | 0.26 |
| SVM | 0.84 | 0.11 | 0.87 | 0.13 | 0.85 | 0.20 |
| Random Forest | 0.84 | 0.19 | 0.82 | 0.12 | 0.88 | 0.27 |
| Gaussian NB | 0.83 | 0.10 | 0.80 | 0.05 | 0.88 | 0.18 |
| Gradient Boost | 0.80 | 0.19 | 0.80 | 0.13 | 0.81 | 0.25 |
| Light GBM | 0.77 | 0.27 | 0.78 | 0.18 | 0.81 | 0.35 |
| XG Boost | 0.72 | 0.41 | 0.67 | 0.39 | 0.80 | 0.45 |
| Decision Tree | 0.62 | 0.35 | 0.63 | 0.38 | 0.62 | 0.35 |

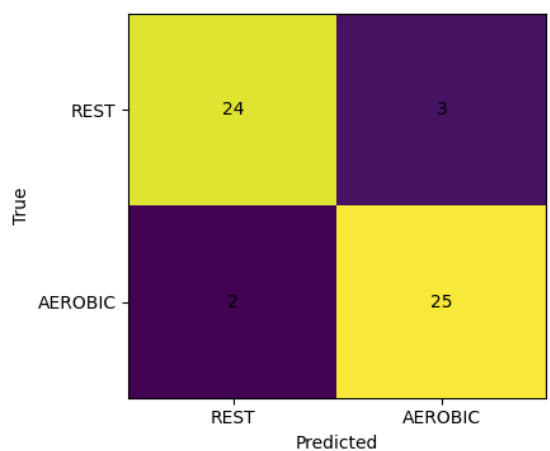

Fig. 4.2 Confusion Matrix for kNN model in 5-fold validation

*B. Model Explainability*

Model Interpretability in this study was addressed through physiologically grounded feature design, conservative feature selection, and statistical analysis. All extracted features were explicitly derived from tonic, phasic, and spectral components of EDA, enabling direct physiological interpretation of model behavior. Feature relevance was assessed using RFE within a linear modeling framework. Model interpretability studies such as SHAP analysis [30] were not applied as the primary aim was to establish conservative subject-independent performance benchmarks.

To further examine feature behavior across conditions, box plot analysis was used to visualize the distribution of selected features across rest and aerobic states. These visualizations showed consistent shifts in magnitude and variability for key tonic and phasic features during aerobic exercise, reflecting increased sympathetic activation.

The physiologically interpretable EDA features exhibiting class-wise separation are shown in Figure 2.5 Phasic amplitude features (e.g., Peak amplitude) and event timing (e.g., SCR peaks per minute) contributed the most discriminative information. These feature importance patterns should be interpreted as indicative of general trends rather than definitive rankings, given the global feature selection strategy and subject-independent evaluation.

IV. DISCUSSION

This study demonstrates that EDA, when conservatively filtered and paired with targeted feature extraction, can distinguish between rest and aerobic exercise under subject-independent evaluation [7], [8]. Using a publicly available dataset, a fixed protocol-based segmentation strategy was applied to isolate rest and aerobic exercise windows. After data quality validation, twenty-three time-domain, tonic, phasic, and frequency-domain EDA features were extracted and systematically evaluated.

Several candidate features, including Mean raw EDA, Mean phasic EDA, Tonic and Phasic variance, Tonic ratio up, and Activation duration, exhibited near-zero values or protocol-dependent behavior across subjects due to normalization and fixed windowing [12], [31], [32]. These features were excluded to reduce sparsity, prevent numerical instability, and avoid introducing bias tied to the experimental structure rather than physiological signal content.

Both correlation analysis and RFE ranking were applied to enhance model stability and interpretability while reducing the risk of overfitting under subject-independent validation. Correlation analysis was first applied to remove redundant features, including Peak density, SCR Interval Mean, SCR Interval RMSSD, 2$^{nd}$ Amplitude, Frequency 2$^{nd}$ Amplitude, and PSD Band power. RFE further refined the feature set by removing SCR Interval Max and Freq Peak Amplitude. The final set of eight features captured complementary aspects of sympathetic sudomotor activity, including sustained low-frequency activation (Mean tonic EDA), signal distribution (Kurtosis, Skewness, and IQR) and phasic dynamics (Mean rise time, Mean recovery time, Peak amplitude, Peaks per minute).

Observed performance across models indicate that EDA alone captures limited but reproducible information about aerobic activity, with LOSO validation highlighting meaningful inter-subject variability rather than model instability [32].

The performance of LDA and kNN indicates that rest and aerobic exercise states can be effectively separated using simple, physiologically interpretable EDA features, without requiring high-capacity nonlinear models. Under subject-independent evaluation, LDA consistently exhibited high sensitivity to aerobic exercise suggesting potential utility as a conservative screening or contextual signal in health-related or safety-critical applications where minimizing missed detections is prioritized. However, the substantial inter-subject variability observed under LOSO suggests that downstream calibration or multimodal integration would be required for reliable deployment across individuals.

From a modeling perspective, no single classifier family dominated across evaluation protocols, indicating that the engineered EDA feature space can support multiple decision paradigms. Among other linear models, GNB shows comparatively weaker performance under LOSO, while LR and kNN performed more robustly. The MLP model demonstrated moderate performance under LOSO, indicating that a shallow neural network architecture can capture meaningful nonlinear structure in EDA features despite limited sample size. Kernel-based methods like SVM demonstrated consistently competitive performance across both LOSO and stratified 5-fold validation suggesting that carefully regularized nonlinear models can generalize with reasonable consistency when paired with conservative feature engineering.

Collectively, these results highlight the trade-off between model complexity and generalization in small biomedical datasets and reinforce the value of selecting moderate-complexity models for wearable physiology applications.

Importantly, these findings support the use of EDA as a contextual signal in early-stage wearable system designs, particularly in scenarios where sensing constraints limit the reliability of motion or cardiovascular signals. Potential applications include monitoring training load in endurance athletes, distinguishing prolonged physical exertion from acute stress in safety-critical occupations such as firefighting or emergency response [9], [10] and providing aerobic-aware rules tied to glucose-uptake for safer insulin dosing strategies in artificial pancreas systems [1], [4], [16]. In such applications, EDA is not intended to replace multimodal sensing but supply complementary sympathetic activation to improve robustness and situational awareness.

Overall, this work provides evidence that carefully processed EDA signals can serve as a viable, non-invasive input for activity-state detection in wearable health systems, while emphasizing the need for subject-independent validation and conservative interpretation when translating wearable bio-signals to real-world use.

## V. CONCLUSION

### A. Limitations

This study is limited by the small number of subjects, which constrained statistical power and necessitated conservative model selection. Exercise and rest segments were defined using fixed protocol-based time windows rather than automatically detected transitions. Because windows are protocol-based, some discriminative power may reflect time-dependent sweating and thermoregulation rather than workload alone [33]. Additionally, only a single aerobic intensity was evaluated, and model performance across varying aerobic workloads was not assessed, limiting generalization to free-living conditions.

Because EDA signals were normalized prior to extraction, tonic-related features reflect relative, low-frequency activation patterns rather than absolute skin conductance level. Thus, caution is advised when extending these findings to clinical or free-living contexts that rely on absolute skin conductance magnitude. Phasic features derived from the normalized signal, such as peak amplitude, represent relative prominence within the segment-specific dynamic range rather than absolute sympathetic output. Accordingly, amplitude-based features should be interpreted as scale-invariant descriptors of activation structure. Timing-based phasic features like phasic mean recovery time remain more directly comparable across conditions.

Feature selection was performed globally rather than within nested cross-validation, prioritizing interpretability over feature stability estimates. In addition, while tonic sequence analysis was initially explored to understand how sympathetic activation evolves over time, the sequence-based features (e.g. tonic slope, activation duration, ratios) were excluded from the final analysis because they increased the risk of overfitting under the fixed windowing protocol [24], [29].

Model interpretability studies such as SHAP analysis [30] were not applied as the primary aim was to establish conservative subject-independent performance benchmarks. Instead, interpretability was addressed through distribution visualization and summary statistics.

Finally, the primary analysis focused on binary rest-aerobic classification. Extension to multi-class or intensity-aware scenarios was intentionally deferred to isolate EDA as an independent physiological signal relevant to fitness and glucose related state detection. Prior studies have demonstrated that classification performance can be improved with the integration of complementary wearable modalities such as heart rate or accelerometry [11], [15].

### B. Future Work

Future work will extend this framework beyond binary rest-aerobic exercise to a three-class classification problem distinguishing stress, aerobic, anaerobic activity. Preliminary results for this extension are included in Appendix B. These results address the limitation of fixed aerobic intensity and enable finer-grained characterization of sympathetic patterns associated with physiologically distinct workloads. Such differentiation is particularly relevant in active occupational settings, where elevated EDA activity might arise from either physical exertion or cognitive stress, and accurate context discrimination is critical for safety and workload assessment [10], [34].

In addition, the proposed EDA-based approach will be evaluated with synchronized glycemic measurements to enable joint analysis of electrodermal activity and continuous glucose measurements. This integration will support investigation of sympathetic activation during physical and cognitive loads, how they modulate glucose trajectories, inform activity-aware safety constraints for closed-loop insulin delivery, and improve hypoglycemia risk prediction in artificial pancreas systems [35], [36].

While multimodal sensing may further increase performance, such extensions were not prioritized as the main objective of this study was to isolate the contribution of EDA and provide a principled foundation for subject-independent sympathetic state awareness in wearable health and safety-critical applications.

## VI. Appendix

### A. Stress vs. Rest (EDA-only) Classification Pipeline

Stress vs. Rest classification using EDA has been extensively demonstrated in prior work and is therefore not presented here as a novel contribution [9], [12], [18]. Rather, this analysis serves as a validation of the preprocessing and feature-engineering pipeline used in the present study and provides a physiological reference point for interpreting performance on the more challenging rest vs. aerobic discrimination task.

To evaluate EDA-only discrimination of stress vs rest, a parallel analysis pipeline was implemented with the same preprocessing and feature engineering methodology, with modifications limited to window selection and class labeling.

For each subject in the stress dataset folder of the publicly available wearable dataset [1], two labeled segments were extracted from the same recording session: Rest segment corresponding to the first 120 seconds of the recording (0-120s), and the Stress segment extracted from a protocol-defined window spanning 630-750 seconds of the recording.

Feature selection was performed using correlation analysis followed by RFE using an LR base estimator. The top ranked features selected by RFE were: Mean Tonic EDA, SCR Interval Max, Kurtosis, Skewness, SCR Peaks per min, Peak Amplitude, Freq Peak Amplitude and PSD Band power.

The same set of benchmark classifiers was trained and evaluated with LOSO validation. Across models, the mean accuracy ranged from 0.83 to 0.91, with corresponding F1 scores between 0.79 to 0.91. Gaussian NB achieved the highest sensitivity to stress with a mean F1 score of 91% (0.91 +/- 0.21) and recall of 97% (0.97 +/- 0.17). Logistic Regression and Extra Trees achieved the highest mean accuracy of 91% (0.91 +/- 0.19).

Performance variability across folds (std. dev. between 0.19-0.30) reflects inter-subject differences in electrodermal responsiveness rather than systematic model overfitting. Overall, these results indicate that physiologically interpretable EDA features are sufficient for subject-independent stress discrimination without requiring high-capacity nonlinear models.

### B. Three-class (Stress-Aerobic-Anaerobic) EDA-only Classification Pipeline

To extend the binary analyses, a three-class classification pipeline was implemented to evaluate whether EDA can discriminate among stress, aerobic, and anaerobic states. The pipeline followed the same preprocessing and feature-engineering framework as the main study, with the expanded label set and analysis applied to the full recordings.

EDA recordings were loaded from three class-specific folders within the wearable dataset, with each file corresponding to a single subject session, under a single condition. Class labels were encoded as Stress=0, Aerobic=1, Anaerobic=2. Tonic, phasic, and frequency-domain features were computed within the feature extraction function using standardized Butterworth filtering operations applied directly to each input segment.

Two additional features were defined: SCR Burst fraction, which quantifies the degree of temporal clustering of phasic responses common in anaerobic activity, and tonic window ramp computed from monotonic increases in tonic trajectory across consecutive 120s windows.

Following correlation analysis and RFE-ranking, the top ten features used in model development were: tonic variance, phasic variance, phasic mean risetime, SCR interval max, SCR interval RMSSD, SCR Peaks per min, IQR, SCR Burst fraction, Freq Peak Amplitude, and Tonic Win Ramp

Benchmark classifiers were trained and evaluated with LOSO validation. LDA achieved the highest overall performance (0.70 +/- 0.32 accuracy, 0.64 +/- 0.36 F1, and 0.70 +/- 0.32 recall). False negative rates (FNR) ranged from 0.27 to 0.42 across the models, indicating systematic class overlaps especially between aerobic vs anaerobic and stress vs anaerobic.

Collectively, these results suggest that EDA alone supports coarse multi-state classification but is insufficient for reliable fine-grained intensity classification without personalization or multimodal integration.

Fig. 6.1 Confusion Matrix for Linear Discriminant Analysis (LDA) model in LOSO validation (3-way classification)

## VII. Author Contributions

[1] Rena Krishna : Conceptualization, data analysis, ML modeling, manuscript drafting

[2] Ramya Sankar : Technology review, methodological guidance, manuscript review

[3] Shadi Ghiasi : Scientific advising, domain experience, manuscript review